\documentclass[a4paper]{article}

\usepackage{INTERSPEECH2019}
\usepackage{multirow}
\usepackage[table,xcdraw]{xcolor}
\usepackage{subfigure}

\title{Audio Source Separation via Multi-Scale Learning with Dilated Dense U-Nets }
\name{Vivek Sivaraman Narayanaswamy$^{\dagger *}$\thanks{$^{*}$The first two authors contributed equally.
This work was supported in part by the ASU SenSIP Center, Arizona State University. Portions of this work was performed under the auspices of the U.S. Depart-ment of Energy by Lawrence Livermore National Laboratory underContract DE- AC52-07NA27344.}, Sameeksha Katoch$^{\dagger *}$, Jayaraman J. Thiagarajan$^{\ddagger}$, \\ Huan Song$^+$ and Andreas Spanias$^{\dagger}$}

\address{$^{\dagger}$Arizona State University, 
  $^{\ddagger}$Lawrence Livermore National Labs,
  $^+$Bosch Research North America
  }
\email{vnaray29@asu.edu, skatoch1@asu.edu, jjayaram@llnl.gov, huan.song@us.bosch.com, spanias@asu.edu}

\begin{document}

\maketitle
\begin{abstract}
  Modern audio source separation techniques rely on optimizing sequence model architectures such as, 1D-CNNs, on mixture recordings to generalize well to unseen mixtures. Specifically, recent focus is on time-domain based architectures such as Wave-U-Net which exploit temporal context by extracting multi-scale features. However, the optimality of the feature extraction process in these architectures has not been well investigated. In this paper, we examine and recommend critical architectural changes that forge an optimal multi-scale feature extraction process. To this end, we replace regular $1-$D convolutions with adaptive dilated convolutions that have innate capability of capturing increased context by using large temporal receptive fields. We also investigate the impact of dense connections on the extraction process that encourage feature reuse and better gradient flow. The dense connections between the downsampling and upsampling paths of a U-Net architecture capture multi-resolution information leading to improved temporal modelling. We evaluate the proposed approaches on the MUSDB test dataset. In addition to providing an improved performance over the state-of-the-art, we also provide insights on the impact of different architectural choices on complex data-driven solutions for source separation.  
\end{abstract}


\section{Introduction}
Audio source separation refers to the problem of extracting constituent sound sources from a given audio mixture. Despite being a critical component of several audio enhancement and retrieval systems~\cite{spanias2006audio}, the task of source separation is severely challenged in practice due to variabilities in acoustic conditions. Mathematically, this is posed as an inverse problem, and classical regularized optimization techniques such as independent component analysis (ICA)~\cite{makino} and matrix factorization are often employed~\cite{thiagarajan2013mixing}. However, such  unsupervised approaches are known to be effective only under specific conditions (e.g. fully determined) and hence several state-of-the-art solutions~~\cite{grais2018raw},~\cite{pascual2017segan},~\cite{stoller2018wave},~\cite{jansson2017singing} increasingly rely on supervisory deep learning techniques, that directly learn the inverse mapping using \textit{mixture-source} pairs. This was motivated by the success of deep learning in solving several highly ill-conditioned inverse tasks in computer vision, such as image completion and super-resolution~\cite{ulyanov2018deep}. A recurring idea in the broad class of recent source separation techniques is to adopt an \textit{encoder-decoder} style architecture, powered by convolutional or generative adversarial networks, for end-to-end optimization of the inversion process. While these data-driven solutions have produced unprecedented success in audio source separation, their performance depends heavily on the choice of data processing strategies and network architectures. 

Until recently, majority of source separation techniques operated in the spectral domain, in particular based on the magnitude spectra. However, by ignoring the crucial phase information, these methods required extensive tuning of the front-end spectral transformation for producing accurate separation results. Recently, in~\cite{stoller2018wave}, Stoller~\textit{et. al.} argued that the need for optimizing spectral transformations can be entirely eliminated by directly operating in the time domain, and that the source recovery quality can be significantly improved by not rejecting the phase information. On the other hand, such a fully time-domain approach necessitates the need to deal with very long temporal contexts at high sampling rates, thus making the network training quite challenging. Stoller~\textit{et. al.} addressed this critical limitation by proposing the \textit{Wave-U-Net} model that leverages multi-scale features obtained using a combination of $1-$D-convolutions and resampling strategies in a U-Net, which is a fully convolutional network widely adopted in semantic segmentation~\cite{ronneberger2015u}. In general, U-Nets are comprised of a \textit{downstream} and an \textit{upstream} module, wherein the former module produces multi-scale features by successively downsampling the audio signals while the latter utilizes resampling in order to produce appropriate context information for subsequent layers. In order to obtain meaningful gradients at different temporal scales, the network allows information propagation between the downstream and upstream layers using skip connections. Though Wave-U-Net outperformed several existing baselines, the optimality of the multi-scale feature extraction process has not been studied yet. Further, conventional upsampling was found to produce undesirable aliasing artifacts, thus requiring the design of an adaptive interpolation scheme.

\noindent \textbf{Proposed Work:} In this paper, we propose crucial architectural changes to the Wave-U-Net model to improve performance of time-domain based source separation systems. First, in lieu of carefully designed resampling schemes, we advocate the use of dilated convolutions to obtain robust multi-scale features. Dilated convolutions have been widely adopted in a variety of sequence modeling tasks~\cite{oord2016wavenet} and are known to be effective in capturing temporal dependencies without any additional parameterizations to standard $1-$D convolutions. More specifically, we show that an adaptive strategy that incrementally leverages information from larger temporal contexts can produce highly robust features. Second, we propose the use of \textit{dense} connections to improve the flexibility in exploiting long-range dependencies. In computer vision applications, DenseNets~\cite{huang2017densely} have produced state-of-the-art recognition performance, through dense connections between convolution layers within a block. While this design was primarily motivated from the standpoint of feature reuse and combating the vanishing gradient problems in very deep networks, we show that this can enable sophisticated modeling of temporal dependencies. 

Using the publicly available MUSDB18 dataset~\cite{musdb18}, we investigate the effects of these architectural changes on the ill-posed inverse task of audio source separation. Following the setup in ~\cite{stoller2018wave}, we assume that the mixing process and the number of sources are known \textit{a priori}. We first examine the impact of the proposed adaptive dilation scheme against other alternative design choices. Subsequently, we show that using dense connections can be particularly beneficial when the depth of the network increases. Our experiments show that the proposed approach, which combines both adaptive dilation and dense connections, significantly outperforms the state-of-the-art baseline.





\section{Related Work}
In this section, we briefly review existing approaches in the literature that utilize deep neural networks for audio source separation. There exists a large body of prior work for source separation using time-frequency representations typically, short-time Fourier transforms (STFTs)~\cite{uhlich2017improving, liutkus20172016, luo2017deep}. While~\cite{uhlich2017improving} and~\cite{liutkus20172016} operated with spatial covariance matrices for source separation in the STFT domain, Luo~\textit{et al.}~\cite{luo2017deep} used the magnitude spectrogram as the representation for a mixture and its constituent sources. Due to inherent challenges in phase spectrum modification, much of existing literature has focused on the magnitude spectrum, while including an additional step for incorporating the phase information, which often leads to inaccurate determination of source signals~\cite{stoller2018wave}. Furthermore, with low-latency systems, large window lengths are needed for effective separation in the STFT domain. 

A common approach to address these drawbacks is to entirely dispense the spectral transformation step and build the estimation algorithm in the time-domain directly. Popular instantiations of this idea include the MultiResolution Convolutional Auto-Encoder (MRCAE)~\cite{grais2018raw}, TasNet~\cite{luo2018tasnet} and the Wave-U-Net~\cite{stoller2018wave}. MRCAE~\cite{grais2018raw} is an autoencoder-style architecture comprised of multiple convolution and transpose convolution layers, wherein each layer supports filters of different sizes. Note that, this is analogous to capturing audio frequencies with multi-scale resolutions. A crucial limitation of this approach is its inability to deal with long temporal sequences - results reported were with $1024$-length sequences, which is often insufficient to model the complex dependencies at high sampling rates. On the other hand, TasNet~\cite{luo2018tasnet}, which is also an encoder-decoder style framework, represents an audio mixture as a weighted sum of basis signals, wherein the estimated weights indicate the contribution of each source signal and the filters from the decoder form the basis set. However, given that the architecture is designed for low-latency scenarios, similar to MRCAE, it deals with only short sequences.

In order to support the use of long temporal sequences, Stoller~\textit{et al.}~\cite{stoller2018wave} proposed the Wave-U-Net model, which uses a U-Net based architecture and can deal with even $80,000$-sample long sequences. While the contracting \textit{downstream} part captures features at different scales, the expanding \textit{upstream} part successively produces high-resolution features. Furthermore, skip connections are used between downstream and upstream layers, in order to obtain meaningful gradients at different temporal scales. However, as we show in this paper, the design of the multi-layer feature extraction process plays a critical role in the performance of this architecture. Furthermore, the training of such multi-scale feature learning networks, particular with very deep \textit{downstream} and \textit{upstream} modules, can be significantly challenging. We propose to incorporate dense connections, that are known to implicitly encourage feature-use~\cite{huang2017densely}, to alleviate this challenge. 

%
\section{Proposed Approach}
The task of audio source separation involves separating a given mixture waveform $M \in \mathbb{R}^{L_m \times C}$ into $K$ constituent source waveforms $[S_{1},\dots S_{K}]$, where each $S_{i} \in \mathbb{R}^{L_n \times C}$. Here, $L_m$ and $L_n$ denote the lengths of the mixture and the sources respectively, and $C$ represents the number of channels. In our formulation, we consider $L_m = L_n$, $C = 2$ implying stereo and the mixing process is a unweighted sum of sources. 

\subsection{Background: The Wave-U-Net Model}
The proposed approach is based on the recent Wave-U-Net architecture in~\cite{stoller2018wave}, which utilizes an \textit{encoder-decoder} style architecture. This model follows a standard U-Net design and is comprised of $12$ convolutional layers, in both the \textit{downstream} and \textit{upstream} parts. Each convolutional layer is followed by a factor $2$ decimation to obtain successively higher resolution information along the \textit{downstream} path. Similarly, in the \textit{upstream path}, bilinear interpolation coupled with an $1-$D convolution layer is used to perform upsampling. In addition, skip connections are included between every convolutional layer in the downstream and upstream paths. 

The number of filters in the first \textit{downstream} layer is fixed at $f = 15$ and is increased in the subsequent layers as $f + f \times (i-1)i$ where $i$ represents the layer index. The kernel size for the filters was chosen to be $15$ in all layers. The \textit{upstream} path also has similar filter configurations except that the kernel size was chosen to be $5$. Finally, the model contains a bottleneck block consisting of a convolution layer with $f + f\times (i)$ filters and kernel size $15$. Note that all convolutional layers included the LeakyReLU activation function. The final source prediction layer uses the \textit{tanh} activation. The loss function for training the model includes the Mean Squared Error (MSE) for each of the sources. Furthermore, an energy conservation constraint is imposed by directly estimating only $K-1$ sources and obtaining the $K^{\text{th}}$ source as the difference between the input mixture and the sum of estimates for $K-1$ sources.

\subsection{Dilated U-Net}
\begin{figure}[t]
	\centering
	\includegraphics[width=\columnwidth]{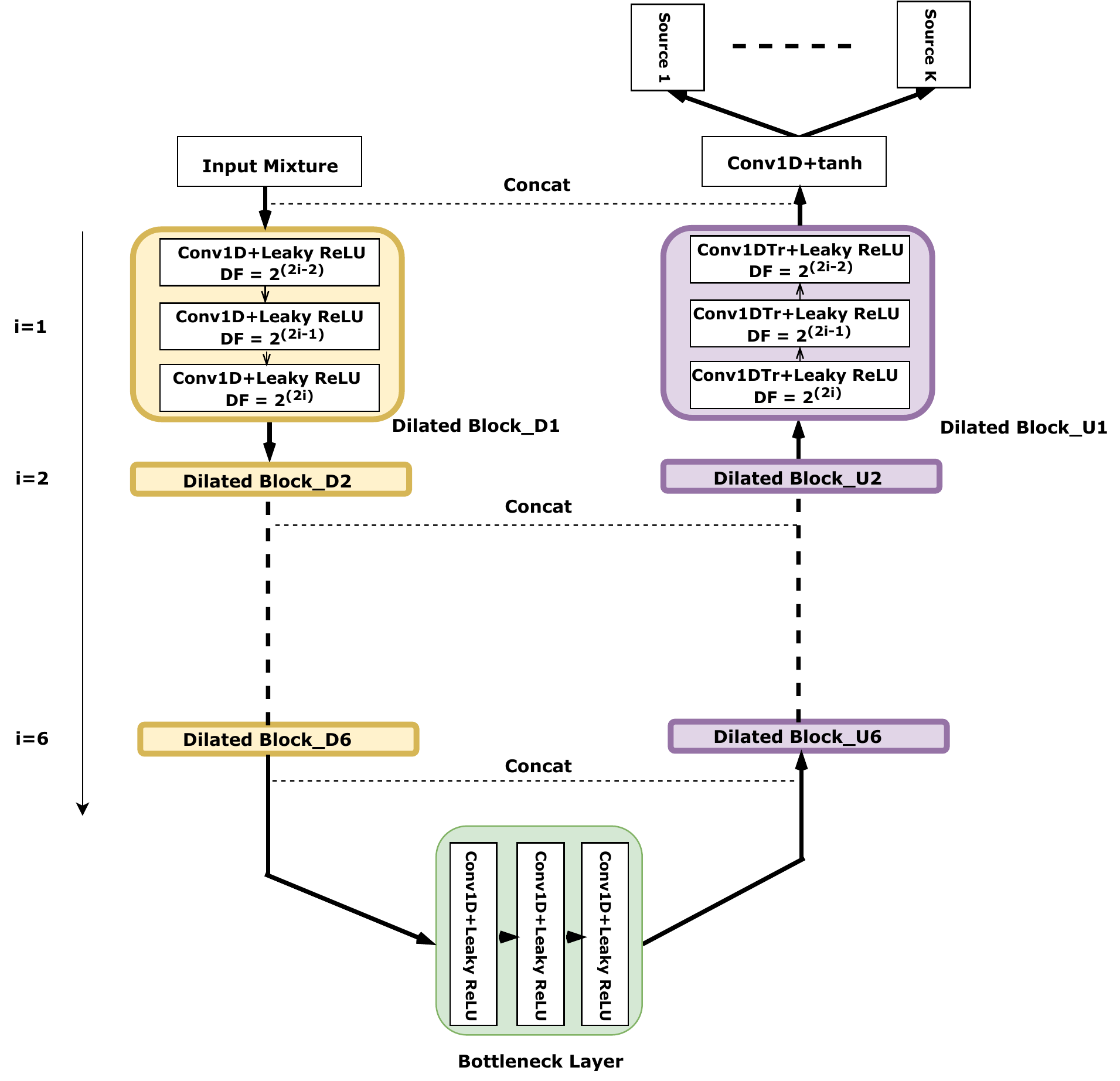}
	\vspace{-2mm}
	\caption{\textit{Dilated U-Net} - Each convolutional block consists of three $1-$D convolutions with exponentially increasing dilation factors. Note that, the \textit{upstream} part utilizes dilated transposed convolutions to recover the sources.}
	\label{fig:dilatedunet}
\end{figure}

As discussed earlier, the performance of source separation approaches that operate directly in the time-domain rely heavily on the quality of the feature extraction process. In particular, building a generalizable model requires the ability to model a wide-range of temporal dependencies, which in turn requires effective multi-scale feature extraction. Furthermore, it was found in~\cite{stoller2018wave} that the choice of resampling scheme was very sensitive. Hence, we propose to employ dilated convolutions to seamlessly incorporate multi-scale features, thereby dispensing the need for explicit resampling. The proposed Dilated U-Net architecture is illustrated in Figure.~\ref{fig:dilatedunet}. 

This model consists of $6$ convolutional blocks in the \textit{downstream} path, where every block contains $3$ dilated convolutions with filter configurations similar to that of Wave-U-Net. Within each block, the dilation rate of the layers increases exponentially by a factor of $2$. We have chosen the dilation rate of the first layer in the consecutive block to be the same as the dilation rate of the last layer in preceding block. This strategy results in providing a wide range of dilation rates from $[1 \dots 4096]$ which increases the effective receptive field, thereby producing improved multi-scale features from the audio excerpt. Note that, all layers perform convolution with a stride of $1$ and employ same padding. The bottleneck block consists of three $1$D convolution layers with dilation rate $1$, stride $1$ and same padding. Correspondingly, the upstream path also consists of $6$ blocks of transposed dilated convolutions, wherein the configurations were chosen to reflect the \textit{downstream} path. The use of skip connections, and the process of source estimation follow~\cite{stoller2018wave}. By retaining the training protocol and loss functions, we hope to quantify the impact of the proposed architectural changes.

\subsection{Dilated Dense U-Net}
\begin{figure}[t]
	\centering
	\includegraphics[width=\columnwidth]{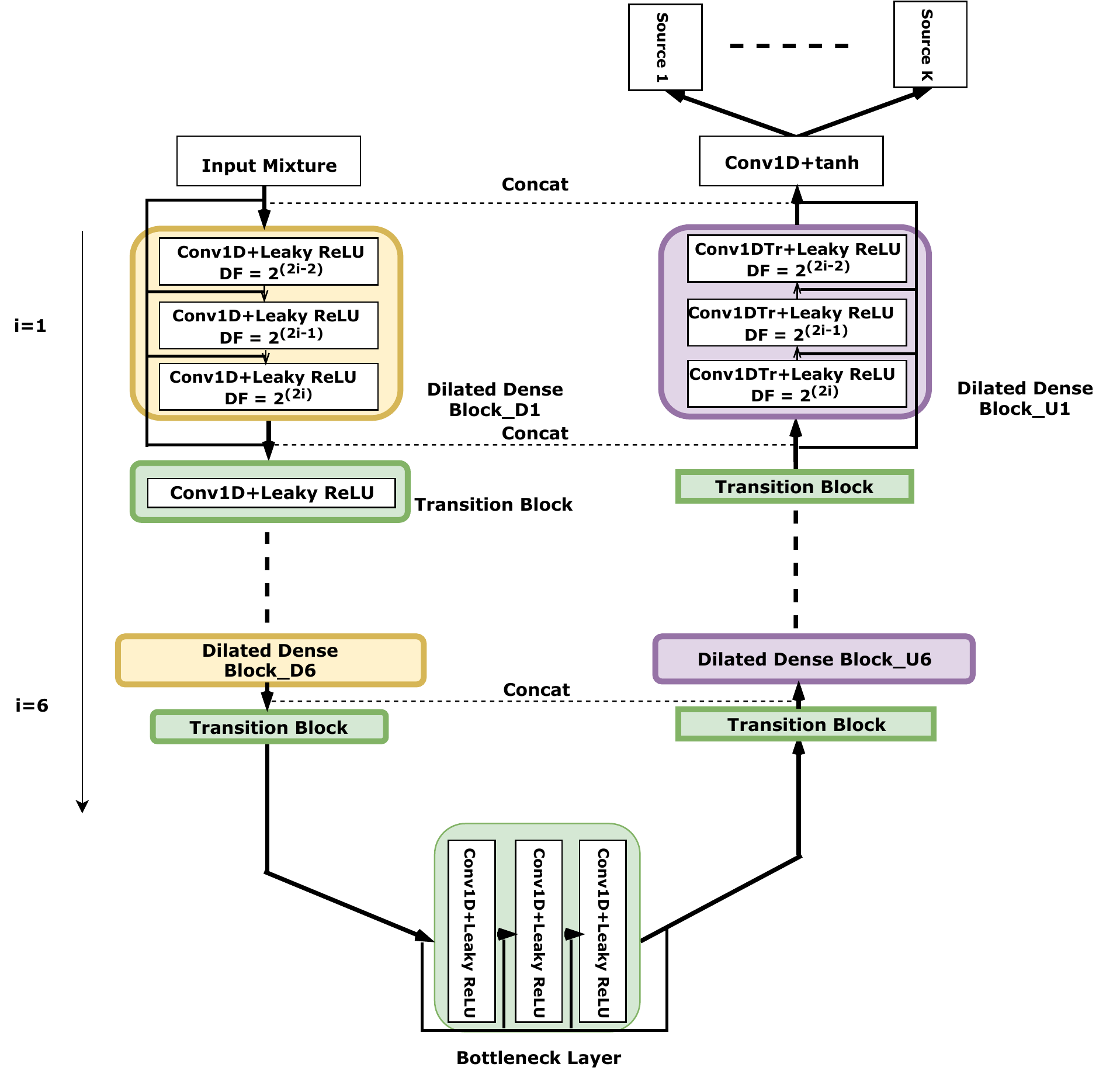}
	\vspace{-4.3mm}
	\caption{\textit{Dilated Dense U-Net} - Similar to Figure \ref{fig:dilatedunet}, every convolutional block is comprised of three $1-D$ convolutions with exponentially increasing dilation factors. In addition, we allow dense connections between convolutions within each block as well as across the \textit{downstream} and \textit{upstream} paths.}
	\label{fig:densedilatedunet}
\end{figure}

While the Dilated U-Net enables seamless incorporation of multi-scale features, with increasing depths in \textit{downstream} and \textit{upstream} paths, the network training becomes very challenging. We propose to improve this by employing dense connections in the networks, that supports feature reuse and protects against vanishing gradients. The Dilated Dense U-Net architecture proposed in this work is illustrated in Figure.~\ref{fig:densedilatedunet}. The architecture is very similar to the previous case, with the key difference that each block (a.k.a dense block), contains dense connections between the dilated convolutional layers. More specifically, within every dense block, the feature maps produced by each layer are concatenated to the subsequent layers in the block to exploit the advantages of feature reuse and improved gradient flow. This can however lead to a large number of feature maps which may be computationally infeasible to process. In order to control the growth of the number of feature maps, we include a transition block which performs dimensionality reduction at the end of every dense block. 

The bottleneck block consists of three $1$D convolution layers that are densely connected with the dilation rates and stride equal to $1$ with same padding. Correspondingly, the \textit{upstream} path consists of $6$ dense blocks where each block contains $3$ transposed convolution layers with dilation rates same as the corresponding block in the \textit{downstream} path. Furthermore, in this model, the skip connections between the respective blocks along the paths are made dense, implying that the feature maps from the block in the \textit{downstream} path are concatenated to all following layers in the corresponding dense block at the upstream path. Finally, the process of source extraction is identical to the Dilated U-Net.
\vspace{-3mm}
\vspace{-0.2mm}
\section{Experiments}
In this section, we evaluate the proposed approaches using the publicly available MUSDB18 dataset and present comparisons to the state-of-the-art Wave-U-Net model~\cite{stoller2018wave}. Before presenting the performance evaluation, we will first discuss the impact of different design choices on the overall performance. This study provides important insights into the behavior of source separation approaches that operate directly in the time-domain.
\begin{figure*}[t]
	\centering
	\subfigure[]{\includegraphics[width=0.48\textwidth]{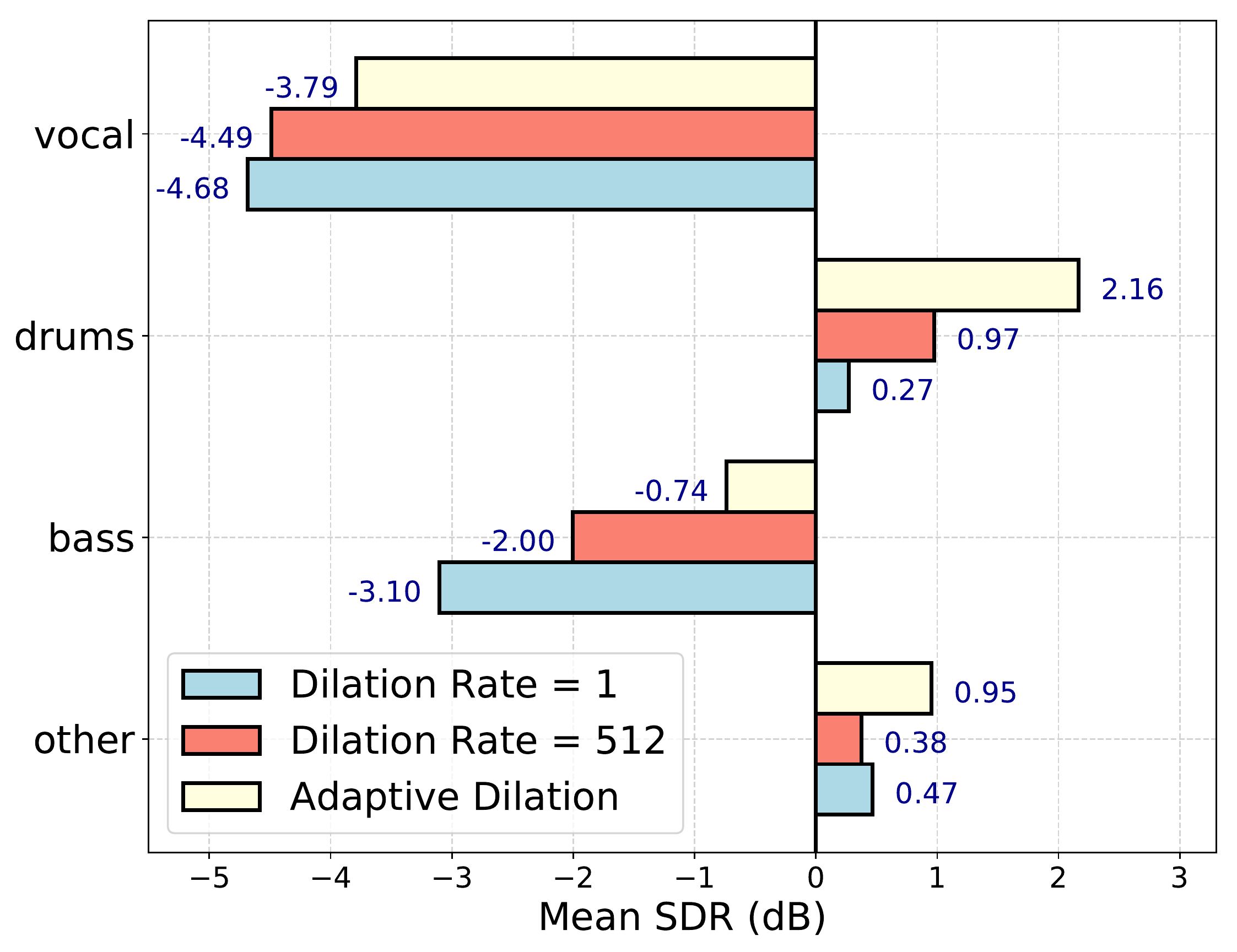} \label{fig:dilated_1_vs_512}}
	\subfigure[]{\includegraphics[width=0.48\textwidth]{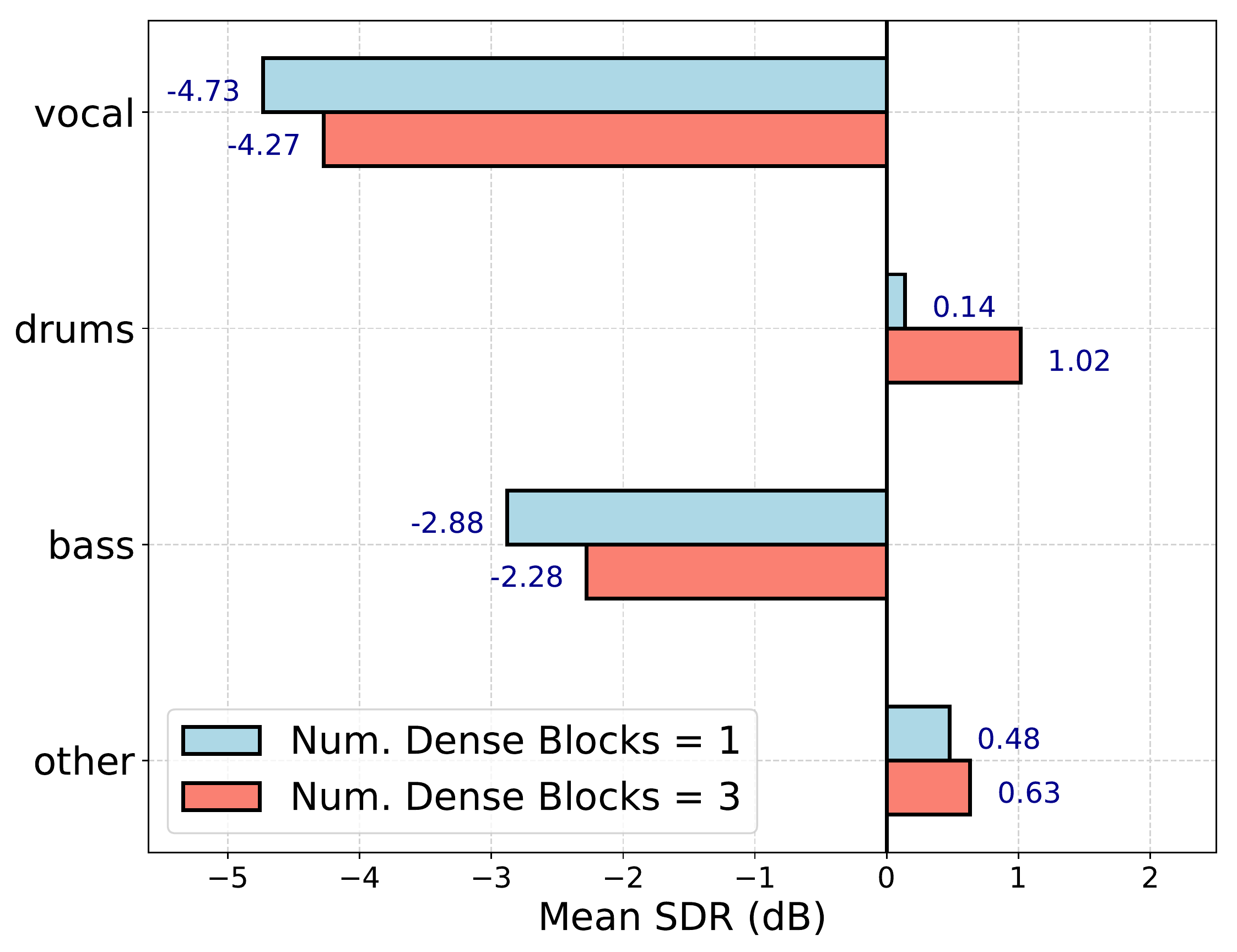}\label{fig:dense_1_vs_2}}
	\vspace{-0.1in}
	\caption{Effect of design choices on the source separation performance (Mean SDR (dB)) - (a) Impact of the choice of dilation rates in different layers of the model. An adaptive learning provides a significant performance boost. (b) Impact of the use of dense connections as the depth of the architecture increases.}
	
\end{figure*}

\begin{table*}[h]
	\centering
	\renewcommand*{\arraystretch}{1.3}
	\caption{Source separation performance obtained using different architectures on the MUSDB18 corpus. We show the mean and median signal-to-distortion ratio (in dB) and in each case the best results are highlighted in bold.}
	\begin{tabular}{|c|c|c|c|c|c|c|}
		\hline
		\multirow{2}{*}{\textbf{Source}} & \multicolumn{2}{c|}{\textbf{Wave-U-Net}}           & \multicolumn{2}{c|}{\textbf{Dilated U-Net}}    & \multicolumn{2}{c|}{\textbf{Dilated Dense U-Net}} \\
		\cline{2-7}
		& \textbf{Mean SDR } & \textbf{Median SDR} & \textbf{Mean SDR} & \textbf{Median SDR} & \textbf{Mean SDR}   & \textbf{Median SDR}  \\
		\hline
		Vocal                                                                            & -3.292                                        & 2.643                                           & -3.787                                        & 2.561                                           & \textbf{-2.986}                                 & \textbf{2.83}                                     \\
		Drums                                                                            & 1.435                                         & 3.310                                           & 2.163                                         & \textbf{3.977}                                  & \textbf{2.449}                                  & \textbf{3.934}                                    \\
		Bass                                                                             & -1.935                                        & 1.942                                           & \textbf{-0.738}                               & \textbf{3.08}                                   & -1.023                                          & 2.711                                             \\
		Other                                                                            & 0.986                                         & 1.911                                           & 0.953                                         & 1.945                                           & \textbf{1.187}                                  & \textbf{2.039}   \\
		
		\hline                                 
	\end{tabular}
	\label{tab:perf}
\end{table*}

\noindent \textbf{Experiment Setup:} We use the MUSDB18 dataset~\cite{musdb18} for our experiments, which is comprised of $75$ tracks for train, $25$ for validation and $50$ for testing. The dataset is encoded in the stems format, and contains multi-stream files of separate sources i.e. bass, drums, other and vocals and resampled to $22050$ Hz. In our experiment setup, we use segments of $16,384$ samples each ($\sim1 sec$) and adopt a simple additive mixing process, following current practice. Note that, in \cite{stoller2018wave}, the authors found that using much larger input contexts ($L_m > L_s$) produces improved results. However, to measure the effective performance of the architectural choices alone, we benchmark without the additional input context. We also performed data augmentation similar to~\cite{stoller2018wave}, wherein the source signals are scaled using a randomly chosen factor in the interval [$0.7$, $1$]. All models reported in the paper were trained using the Adam optimizer with a learning rate of $0.0001$ and a batch size of $16$. While the results for the initial study were obtained by training for only $30$ epochs, the actual performance metrics were obtained by training for longer ($\sim80$ epochs). The mean and median signal-to-distortion ratio (SDR) for each of the sources over the entire dataset are computed. The SDR metric takes into account the noise arising from interference and other artifacts in the estimated audio sources~\cite{sdr_vincent}. The mean SDR is computed after removing silence regions. Since the mean value can be affected by outliers from near-silence regions we also report the median SDR which is known to be more unbiased. 



\subsection{Impact of Design Choices}
As discussed earlier, the source separation performance depends heavily on the architecture choices for multi-scale feature extraction. Hence, we first study the impact of different dilation schemes in the proposed architecture, wherein we entirely eliminate the resampling process using dilated convolutions. As described in the previous section, our architecture is comprised of $6$ blocks of convolutional layers. In its simplest form, we use conventional $1-$D convolutions with the dilation rate fixed at $1$ in all layers. In addition, we consider the case where it was fixed at a constant value ($512$) and the case with the proposed adaptive dilation scheme. As observed in Figure \ref{fig:dilated_1_vs_512}, the sub-optimal performance of conventional $1-$D convolution clearly shows the importance of leveraging multi-scale features. Furthermore, the proposed adaptive dilation scheme provides a significant performance boost compared to using fixed dilation in all layers. Similarly, we analyzed the impact of using dense connections on the separation performance. For this experiment, we fixed the dilation rate at a constant value of $512$ and the number of convolutional blocks at $1$ and $3$ respectively. As showed in Figure \ref{fig:dense_1_vs_2}, as the depth of the network increases, using dense connections provides significant gains.


\subsection{Performance Evaluation}
In this section, we report the overall performance of the proposed approaches, namely Dilated U-Net and Dilated Dense U-Net, on the MUSDB18 dataset. Though a number of baseline techniques exist for time-domain source separation, we chose to compare against the state-of-the-art Wave-U-Net architecture from~\cite{stoller2018wave}. Table \ref{tab:perf} compares the mean/median SDR (dB) for each of the constituent sources for the testing set in MUSDB18. The first striking observation is that by improving the multi-scale feature extraction process, we could obtain significant performance improvements over the baseline in all cases. In particular, our approaches provide improvements between $0.2$dB and $1.2$dB. While the dilated variant eliminates the need for explicit resampling by capturing information from exponentially increasing receptive fields, the inclusion of dense connections improves the robustness of the training process. This performance gain clearly evidences the dependence of these complex data-driven solutions for audio processing on the inherent feature extraction mechanism, and the need for improved architecture design.

\bibliographystyle{IEEEtran}
\bibliography{main}

\end{document}